\pdfoutput=1

\documentclass[11pt]{article}

\usepackage[preprint]{acl}

\usepackage{times}
\usepackage{latexsym}

\usepackage[T1]{fontenc}

\usepackage[utf8]{inputenc}

\usepackage{microtype}

\usepackage{inconsolata}

\usepackage{graphicx}

\usepackage[table]{xcolor}
\usepackage{colortbl}
\usepackage{array}
\usepackage{booktabs}
\usepackage[most]{tcolorbox}
 \usepackage{amsmath} 

\title{How Model Size, Temperature, and Prompt Style Affect LLM-Human Assessment Score Alignment
}

\author{
  \textbf{Julie Jung\textsuperscript{1}},
  \textbf{Max Lu \textsuperscript{*}\textsuperscript{1}},
  \textbf{Sina Chole Benker\textsuperscript{2}},
  \textbf{Dogus Darici\textsuperscript{2,3}}
\\
  \textsuperscript{1}Harvard Graduate School of Education,
  \textsuperscript{2}Munster University,\\
  \textsuperscript{3}Institute of Anatomy and Neurobiology, University of Münster
\\
  \small{
  * Joint First Authors
  }
  \\
  \small{
    \textbf{Correspondence:} \href{mailto:maxlu@fas.harvard.edu}{maxlu@fas.harvard.edu} 
  }
}

\begin{document}
\maketitle
\begin{abstract}
We examined how model size, temperature, and prompt style affect Large Language Models' (LLMs) alignment within itself, between models, and with human in assessing clinical reasoning skills. Model size emerged as a key factor in LLM-human score alignment. Study highlights the importance of checking alignments across multiple levels.

\end{abstract}

\section{Introduction}
The rapid advancement of Large Language Models (LLMs) has introduced new possibilities for evaluating text-based and conversational assessments, offering a scalable and cost-effective alternative and complement to traditional human scoring. While earlier approaches relied on statistical models, recent studies have demonstrated that LLMs can produce accurate, consistent, and personalized evaluations, which are particularly valuable in settings with limited access to expert raters and a need for timely feedback (e.g., \citealp{pack2024large}; \citealp{xiao2024automation}). In educational contexts, assessments leveraging automated scoring have enabled more frequent formative and summative evaluations by reducing teacher workload and accelerating feedback cycles (\citealp{bailey2025addressing}). These developments raise critical questions about how such technologies compare to current assessment practices and what implications they hold.

Before LLM-generated scores can be meaningfully deployed, it is essential to consider the impact of their opaque scoring processes and how different settings influence their behavior. Although LLMs may approximate human ratings at the surface level, there is often a misalignment between the cognitive processes humans rely on and the pattern-based inferences LLMs make (\citealp{baldwin2025vulnerability}). Unlike human raters who may draw on domain-specific reasoning, LLMs operate largely as ``black boxes,'' making it difficult to trace how inputs lead to particular scoring outcomes (\citealp{bathaee2017artificial}). This lack of transparency may be problematic in domains where judgments need to be informed by specific expertise. Additionally, LLMs are not a single, homogeneous system. Different models behave differently, and even the same model can produce varying results depending on its settings. As a result, establishing a strong body of validity evidence is critical before integrating LLMs into assessment practices, particularly as a complement to human scoring.

In this study, we investigate how LLM-generated ratings, with varying model size, temperature, and prompt style, compare to current practices in a specific case study of clinical reasoning skill assessments in medical education. Clinical reasoning, the ability to recognize and interpret a patient’s needs and health condition, is an essential skill for all health professionals (\citealp{tanner2006thinking}). It requires problem-solving abilities, experiential knowledge, and deliberate decision-making. Pattern recognition, in particular, is strongly linked to diagnostic success (\citealp{coderre2003diagnostic}), yet many medical students lack the clinical exposure needed to develop this skill. Thus, there is a need to provide students with opportunities to practice encounters with patients and receive structured, targeted feedback to develop their clinical reasoning skills (\citealp{haring2017observable}). However, formative assessment of clinical reasoning is resource-intensive, typically requiring advanced medical experts to review and score student-patient dialogue. While LLMs have been used to simulate patients in such dialogues, further research is needed to determine whether they can be reliably used for scoring, and how different configurations of LLMs affect their scoring behavior and reliability \citep{brugge2024large}. This study explores how LLM-based evaluations align with or diverge from current human-based assessments and what those findings imply for the validity, fairness, and practical use of LLMs in medical education.

\section{Literature}
LLMs have shown great promise in evaluating complex, language-based assessments. Recent studies comparing various LLMs have demonstrated that some models, such as ChatGPT using GPT4, have high alignment with human scorers for tasks such as essay assessment, although they tend to exhibit a tendency toward inflated scores (\citealp{sessler2025can}). LLMs are also capable of extracting nuanced insights from text, aligning reasonably well with human ratings in tasks such as discourse coherence analysis (\citealp{naismith2023automated}) and evaluating short-answer assessments in an undergraduate medical program (\citealp{morjaria2024examining}) with similar performance to human expert assessors. However, most studies have focused on scoring outcomes rather than examining the processes through which LLMs generate their ratings. Given the ``black box'' nature of LLM reasoning (\citealp{bathaee2017artificial}), further research is necessary to better understand how these models arrive at their judgments.

Although directly understanding the internal processes of LLM scoring remains challenging due to their nature, insights can be indirectly gained by systematically exploring how prompt style and model parameters influence scoring outcomes. Morjaria and colleagues (\citeyear{morjaria2024examining}) found that the absence of a rubric in the prompt given to the LLM improved their alignments with human scoring, suggesting prompt style significantly shapes how these models evaluate responses. Yet, existing studies have largely overlooked how prompting the LLM to adopt a certain persona, such as a clinical expert, may further improve its alignment with human expert judgments by capturing nuanced reasoning processes underlying expert scoring. Additionally, model parameters, such as temperature settings, influence the randomness of and variability of generated responses and thus may affect scoring consistency and accuracy. Official guide on OpenAI suggests that lower temperature leads to more deterministic and less random outputs, while higher temperature leads to more random outputs \citep{openai2023createtranslation}. Many studies have linked the randomness to ``creativity'' or variability in the response \citep{peeperkorn2024temperature}. While the specific temperature setting for public-accessible ChatGPT is undisclosed, it is commonly believed to be around 0.7 \citep{uvic2024temperature}. Lastly, the choice of model type or size could also substantially influence scoring alignment due to variations in reasoning capabilities (\citealp{sessler2025can}). As psychometrician Andrew Ho emphasized, it is essential to ask, ``Who is using what score for what purpose?'' Different types of assessments, tailored to different use cases, demand varying standards for scoring, such as consistency and inter-rater reliability (\citealp{ho2025metaphors}). Accordingly, as LLMs are increasingly employed as assessors, it becomes critical to systematically examine how variations in prompts, temperature settings, and model size affect their performance across diverse contexts. Such investigation is a necessary step toward the effective and trustworthy adoption of LLM-generated assessments.

When it comes to using LLMs to evaluate student performance for formative feedback, it is essential to understand how reliable are their ratings—how does LLM ratings compare itself to other LLMs and, importantly, to that of human—and what these comparisons imply for validity and practical use. Clinical reasoning provides a challenging test case in that it encompasses a complex cognitive process that relies on both formal and informal reasoning strategies, deeply informed by domain-specific knowledge (\citealp{simmons2010clinical}). It involves collecting patient information, evaluating its clinical significance, forming inferences, and generating diagnostic hypotheses. To facilitate formative assessment of these skills, the Clinical Reasoning Indicators-History Taking-Scale (CRI-HT-S) was developed, grounded in clinical reasoning indicators identified through qualitative research (\citealp{haring2017observable}). This scale assesses dimensions such as a medical student’s capacity to lead patient conversations and ask questions in a logical sequence. An initial validation has shown that the CRI-HT-S demonstrates acceptable reliability and internal consistency for assessing undergraduate medical students' clinical reasoning skills (\citealp{furstenberg2020assessing}). Given the complexity inherent to clinical reasoning, it is particularly important to explore how LLMs evaluate this nuanced and multidimensional construct.

Despite growing interest in leveraging LLMs for formative evaluations, existing research has not systematically investigated how interactions between critical LLM parameters (model size, temperature, and prompt style) influence their scoring alignment with expert human assessors. Our study addresses this gap by comprehensively evaluating the impact of these parameters within clinical reasoning assessments, providing guidance for valid and practical implementation of LLMs in educational contexts. Specifically, we address the following research questions:

\textbf{RQ1}:How consistent are LLM raters when re-rating the same student in the clinical reasoning assessment?

\textbf{RQ2}: How do LLM design parameters (model size, temperature, and prompt style) affect the inter-rater reliability and alignment between LLM raters in the clinical reasoning assessment? 

\textbf{RQ3}: How do LLM design parameters (model size, temperature, and prompt style) affect the inter-rater reliability and alignment with human raters in the clinical reasoning assessment?

\textbf{RQ4}: How do LLM design parameters (model size, temperature, and prompt style) and their interactions affect the average score levels in the clinical reasoning assessment compared to human raters?

\section{Data and Sample}

Our dataset consists of the transcripts of 21 third-year medical students in Germany who were each engaged in conversations with four fictional patients with differing diagnoses to assess the students’ clinical reasoning skills. Two human raters with medical expertise rated each transcript with the Clinical Reasoning Indictors-History Taking-Scale, which consists of eight items measuring the quality of the medical student’s clinical decision making on a Likert scale from 1 to 5 (see Appendix \ref{sec:appendixA}). We systematically varied three LLM parameters: model size (GPT-4o as the large model vs. GPT-4o-mini as the small model), temperature (low: 0.2 vs. high: 0.7), and prompt style (regular vs. expert prompt; see Appendix \ref{sec:appendixA} \& \ref{sec:appendixB}).These variations resulted in eight distinct LLM configurations. Each model rated every student dialogue, producing eight item-level scores per student. All transcripts were in German. One conversation (student 16 with patient 1) was excluded due to missing dialogue. Students were informed that their interactions would be assessed for clinical reasoning.

\section{Methods}

\begin{table}[ht]
  \centering
  \small 
  \resizebox{\columnwidth}{!}{%
  \begin{tabular}{c c c c}
    \toprule
    \textbf{Rater} & \textbf{Model} & \textbf{Temperature} & \textbf{Prompt Style} \\
    \midrule
    LLM1 & \cellcolor{green!10}Small (gpt-4o-mini) & \cellcolor{orange!20}Low (0.2) & Default \\
    LLM2 & \cellcolor{green!10}Small (gpt-4o-mini) & \cellcolor{orange!20}Low (0.2) & \cellcolor{gray!30}{Expert Persona} \\
    LLM3 & \cellcolor{green!10}Small (gpt-4o-mini) & \cellcolor{orange!40}High (0.7) & Default \\
    LLM4 & \cellcolor{green!10}Small (gpt-4o-mini) & \cellcolor{orange!40}High (0.7) & \cellcolor{gray!30}{Expert Persona} \\
    LLM5 & \cellcolor{green!30}Large (gpt-4o)     & \cellcolor{orange!20}Low (0.2)  & Default \\
    LLM6 & \cellcolor{green!30}Large (gpt-4o)     & \cellcolor{orange!20}Low (0.2)  & \cellcolor{gray!30}{Expert Persona} \\
    LLM7 & \cellcolor{green!30}Large (gpt-4o)     & \cellcolor{orange!40}High (0.7) & Default \\
    LLM8 & \cellcolor{green!30}Large (gpt-4o)     & \cellcolor{orange!40}High (0.7) & \cellcolor{gray!30}{Expert Persona} \\
    \bottomrule
  \end{tabular}
  }
  \caption{Color-coded settings of LLM models by Rater ID. Background colors indicate model size and temperature, with red boxes around prompt style when set to "Pretend to be expert."}
  \label{tab:ai-settings}
\end{table}

To address the first research question, each student’s responses were rated twice by the LLM raters. Each student received scores on eight items across four rounds of conversation. For each rater, we averaged a student’s item scores across all rounds. We then calculated pairwise intraclass correlation coefficients (ICCs) using the icc() function from the irr package in R (\citealp{irr}) across all raters. A two-way model was used to appropriately assess absolute agreement across the two trials of the same rater when students are considered random effects. According to the classical test theory, a high ICC suggests that the LLM tends to agree with its own ratings, a sign of reliability but not necessarily validity. However, a low ICC would suggest poor agreement of the same LLM across trials, a sign of a lack of reliability and validity. Conventionally, ICC below 0.5, between 0.5 and 0.75, and above 0.75 suggests poor, moderate, good or excellent reliability, respectively (\citealp{koo2016guideline}). 

To answer the second and third research questions, we evaluated interrater agreement between LLMs and across human and LLM ratings. To facilitate pairwise comparisons between LLM and human raters, we also created a composite ``averaged human'' rater by averaging the two human scores. We used a two-way model to compute the ICC across all pairs of raters (8 LLM raters, 2 human raters, and 1 ``averaged human'' rater). Comparing ICC values between rater pairs allowed us to assess the degree of alignment between them. A high ICC between LLM and human indicate that the LLM rater may be scoring similarly to humans. Meanwhile, high ICC between LLM raters alone reflects consistency between LLMs but not necessarily alignment with human judgment. 

To address the fourth research question, we further aggregated the item-level scores into a single person-level mean score per student per rater. This approach minimizes the influence of within-person variability and allows us to focus on between-student differences. However, this simplification assumes that the eight items reflect a unidimensional latent construct of performance, which is a limitation of this method.

To examine how different LLM design parameters affect LLM-generated ratings, we constructed a model including main effects and interactions among model size ($large$), temperature setting ($hi\_temp$), and prompt style ($expert$) to reflect the eigh different LLM rater configurations. Each of these factors was coded as a binary indicator. For student $i$ rated by rater $j$, we fit the following model using heteroskedasticity-consistent standard errors:

\begin{align}
\textit{score}_{ij} =\, & \beta_0 + \beta_1 \textit{large}_j + \beta_2 \textit{hi\_temp}_j + \beta_3 \textit{expert}_j \notag \\
& +\, \beta_4 \textit{large}_j \cdot \textit{hi\_temp}_j \notag \\
& +\, \beta_5 \textit{large}_j \cdot \textit{expert}_j \notag \\
& +\, \beta_6 \textit{hi\_temp}_j \cdot \textit{expert}_j \notag \\
& +\, \beta_7 \textit{large}_j \cdot \textit{hi\_temp}_j \cdot \textit{expert}_j 
+ \varepsilon_{ij}
\end{align}
This model estimates not only the average effects of each design parameter but also their two-way and three-way interaction effects, allowing us to explore whether the impact of one parameter depends on the levels of the others.

\section{Results}

\begin{figure}[t]
    \centering
  \includegraphics[width=\linewidth]{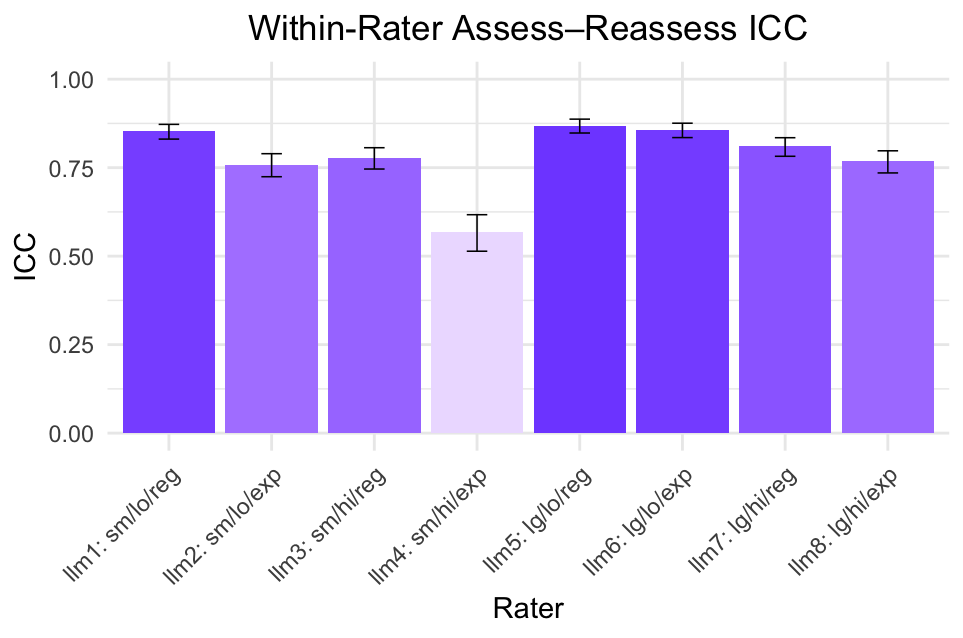}
  \caption{Bar plot of the intraclass correlations (ICCs) between two repeated assessments of the same student for each type of LLM rater. 95\% confidence intervals are included.}
  \label{fig:ratericc}
\end{figure}

For the within model consistency (RQ1), we compared the test–retest ICCs across LLM raters (see Figure \ref{fig:ratericc}). Most models showed high reliability (ICC$\approx$.76–.87). The exception was GPT-4o mini with high temperature and the expert prompt, which exhibited lower consistency. In contrast, GPT-4o with low temperature and the regular prompt achieved the highest reliability (ICC = .87).

\begin{figure*}[h]
    \centering
  \includegraphics[width=\linewidth]{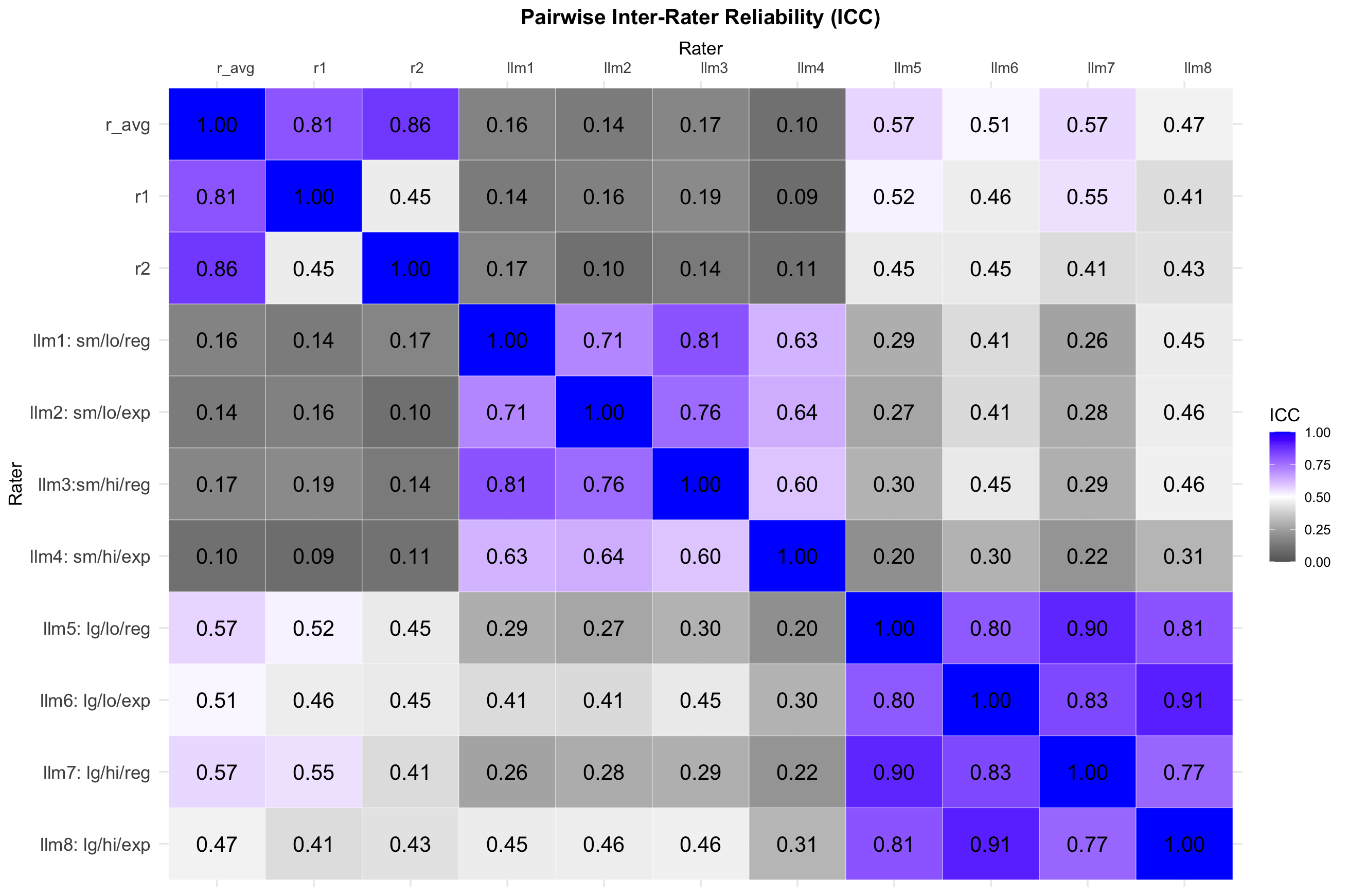}
  \caption{Pairwise intraclass correlation coefficients (ICC) between human and LLM raters across rating conditions.}
  \label{fig:icc}
\end{figure*}

For consistency between models and between human and model, figure~\ref{fig:icc} shows the pairwise intraclass correlation coefficients (ICCs) among all raters. As a reference, the two human raters (r1 and r2) showed moderate agreement (ICC = 0.45). To answer RQ2, which focused on consistency between models, the four small-model LLM raters (llm1–llm4) showed good to very good internal agreement (ICC range: 0.60–0.81), and the large-model LLM raters (llm5–llm8) showed even stronger internal agreement (ICC range: 0.77–0.91).

However, further inspection revealed discrepancies between human and LLM ratings (RQ3). There appeared to be poor agreement between human raters and smaller LLMs (ICC < 0.20). In contrast, there appeared to be moderate agreement between humans and larger LLMs (ICC = 0.41–0.57), comparable to or slightly exceeding human–human agreement. These results indicated that while language models, regardless of size, could be highly agreeable with each other, this does not guarantee alignment with human evaluations. Therefore, if human ratings are considered the benchmark, alignment should be assessed based on direct agreement with human raters, rather than relying solely on internal consistency among the models.

Examining the ICCs across individual items to identify discrepancies between LLM-generated and human scores revealed interesting patterns. For example, item 4, assessing whether the student’s questions suggested specific causes of symptoms, showed high consistency among LLM raters but low alignment between LLM-generated and human scores. This suggested that LLMs and human raters may interpret the criterion of ``suggesting specific causes'' differently, emphasizing the need for improved prompts that more effectively capture human evaluative reasoning (see Appendix \ref{sec:appendixC}).

For RQ4, we estimated a linear model with robust standard errors to examine how model size, temperature, and prompt style, as well as their interactions, affected the average clinical reasoning scores produced by different LLM raters (see Appendix \ref{sec:appendixD}). As shown in Figure \ref{fig:interaction}, all LLM raters, regardless of configuration, produced higher average student scores than the human rater average (M = 3.19, SE = 0.054). The estimated intercept represented the expected score under the baseline configuration (small model, low temperature, regular prompt) and was significantly above the human rater average ($\beta$ = 3.52, p < .001), indicating that this model configuration was less stringent than the human raters.

\begin{figure}[t]
    \centering
  \includegraphics[width=\linewidth]{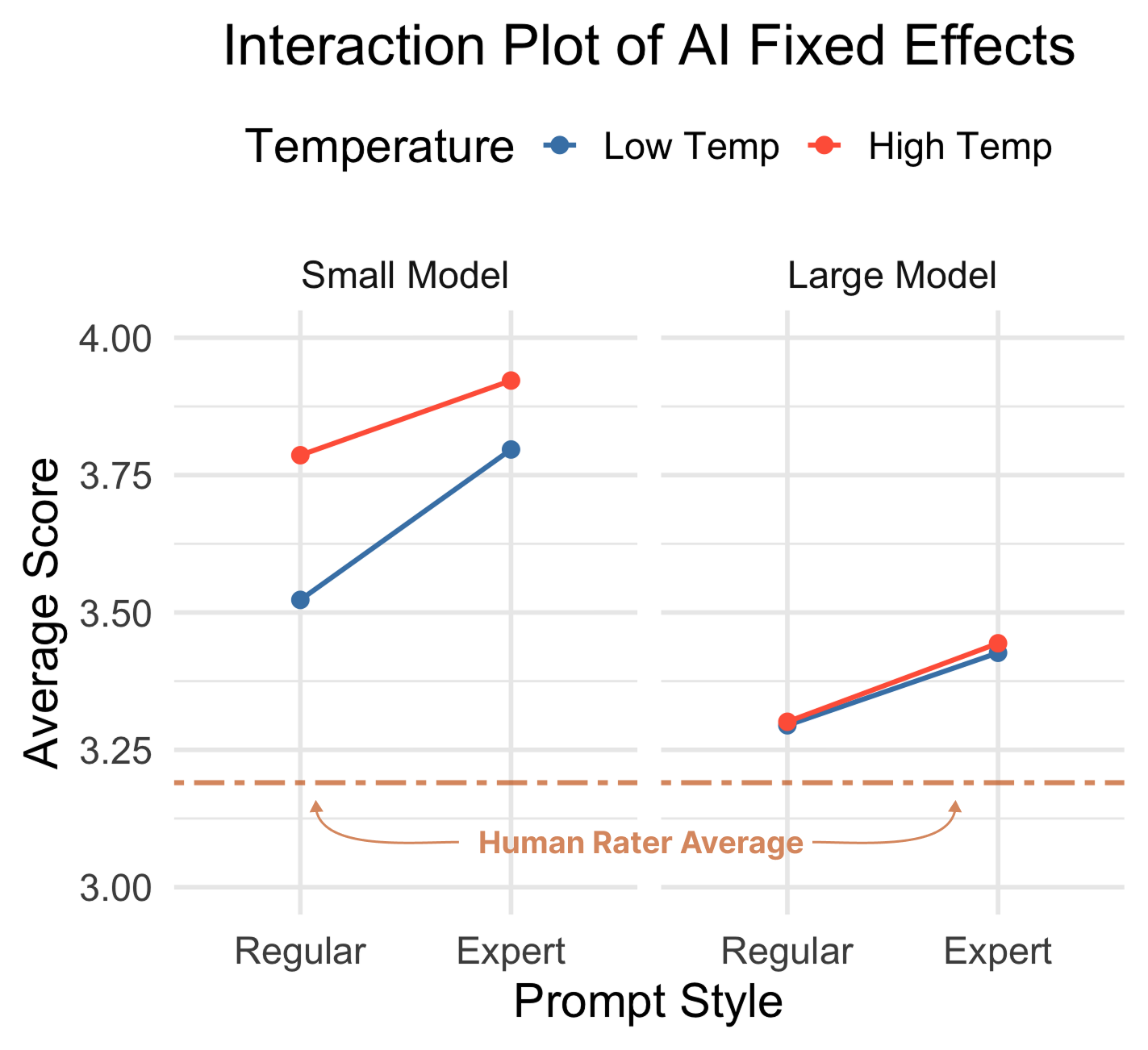}
  \caption{Interaction plot showing how model size, temperature, and prompt style influence AI-generated average scores compared to human rater average.}
  \label{fig:interaction}
\end{figure}

Both high temperature and expert prompting were associated with significantly higher scores overall ($\beta$ = 0.26 and $\beta$ = 0.27, respectively; both p < .01). The interaction between model size and temperature was negative and statistically significant ($\beta$ = –0.26, p < .05), suggesting that higher temperature inflated the score more for the small model than the large one, holding prompt style constant. On the other hand, the interaction between model size and expert prompting was nonsignificant ($\beta$ = –0.14, p > .1), indicating that the effect of expert prompting on average scores did not significantly differ between the large and small models, holding temperature constant. The interaction among all three factors was also nonsignificant ($\beta$ = 0.15, p > .1), suggesting that the combined effect of temperature and prompt style did not differ significantly between large and small models.

For the small model, expert prompt and high temperature each led to increases in average scores, up to almost 3.9. In contrast, the large model showed a smaller range of scores across conditions, around 3.3 to 3.4, regardless of prompt style or temperature. Notably, if the human rater average is taken as the gold standard, the configuration most closely aligned with human scoring norms was the large model with regular prompt and low temperature (M = 3.29, SE = 0.065).

\section{Discussion}
The purpose of this study was to explore how variations in model settings influence LLM-generated ratings and how these scores align within themselves, with each other, and with human scores, given the intended formative use of this assessment. We examined the impact of model size, temperature, and prompt style. Overall, the results showed that certain combinations of these parameters may be more suitable for specific assessment purposes. In evaluating medical students’ clinical reasoning skills, findings revealed that model size influenced both alignment with human raters and internal consistency, with GPT-4o consistently outperforming GPT-4o-mini. Temperature and prompt style had a relatively minor effect when using the larger model compared to the smaller one. Thus, for formative assessment of clinical reasoning, employing a larger model such as GPT-4o is recommended to achieve greater consistency and human alignment.

Our results revealed that LLMs showed high ICC within the same model and between models of the same size, but can systematically differ from human ratings. Across almost all models, the assess–reassess reliability was fairly high, indicating fairly high reliability in reproducing their ratings across different trials. The four smaller models generally have ICC higher than 0.6 and the four larger models have ICC higher than 0.77, suggesting high internal consistency between models with the same size. However, when it comes to agreement with human, their performance vary. The smaller model (GPT-4o mini) showed very poor agreement with human raters, while the larger model showed ICC levels that are comparable to human-human ICC. Combining these insights highlight a key feature or limitation of LLM raters: high internal consistency within or across LLM raters does not imply alignment with human judgments. Simply deploying a group of LLM raters and observing agreement among them is insufficient for validating their use in scoring. Direct comparison with human ratings is necessary in those cases. However, once we are able to configure an LLM that has a high agreement with human, the high assess–reassess consistency is encouraging sign that such an LLM can produce reliable ratings.

In this study, although human raters themselves showed only moderate inter-rater reliability, the larger model aligned more closely with human scores than the smaller models did. Additionally, human raters gave lower average scores than the LLMs, suggesting that humans may apply stricter evaluation criteria - a pattern consistent with prior findings (\citealp{morjaria2024examining}). Therefore, incorporating at least one human rater, or ensuring that LLM raters are demonstrably aligned with human evaluations, remains essential when human judgment serves as the gold standard.

Future implementations should consider periodic human validation of LLM-generated scores to promote fairness and reliability. Importantly, assessments should avoid assigning LLM raters to some students and human raters to others, as this could introduce systematic biases. Further research with larger sample sizes should also explore the use of Generalizability Theory to analyze the impact of rater variability across combinations of human and LLM raters (\citealp{shavelson1992generalizability}).

Another notable finding was related to scores generated by LLMs using persona prompts. Contrary to our expectation that persona prompting (asking LLM to pretend to be an expert) would enhance alignment with human expert ratings, these prompts instead led to higher scores compared to standard prompts and showed poorer alignment with human ratings. This suggests that persona-based prompting, at least in this case, may not effectively replicate human expert evaluation processes. Future research could explore alternative prompting strategies, such as few-shot prompting, or adding a few examples within prompts, as a potentially more effective method to improve alignment between LLM-generated ratings and human ratings, particularly for items exhibiting notable discrepancies, such as Item 4.

Ultimately, advancing the use of LLMs for assessment requires careful attention not only to consistency within models, but also to variation between models and, most critically, to their alignment with human judgments, as overlooking any of these dimensions risks undermining the validity of LLM-generated scores and even leading to systematic biases.

\section*{Limitations}

A major limitation of this study was the relatively small sample size. Having only two human raters restricted our ability to establish a robust human expert benchmark. This could have influenced the reliability comparisons with LLM-generated scores. Additional human raters would reduce measurement error. Moreover, with scores from only 21 medical students, we were unable to comprehensively explore or decompose sources of measurement error contributing to discrepancies between variations of LLM-generated and human ratings. Future studies should include larger datasets, enabling the application of Generalizability Theory to better identify and quantify multiple facets of error, such as variability due to raters, tasks, or specific scoring criteria.

\section*{Acknowledgments}

We thank Prof. Andrew Dean Ho, Kenji Kitamura, and Mikayla Do for their valuable guidance and advice.


\bibliography{main}

\appendix

\section{Regular prompt}
\label{sec:appendixA}

The regular prompt we used is shown below:

\begin{tcolorbox}[title=Regular Prompt, colback=gray!5!white, colframe=black!75!white, sharp corners=southwest, fonttitle=\bfseries]
\small
At this point, you will assess the quality of the third-semester medical student conducting a history-taking conversation.

Your assessment should include the following eight criteria

1. Assess whether the user has taken control of the conversation to obtain the necessary information.

2. Assess whether the user recognizes all relevant information.

3. Assess whether the user formulates targeted questions so that he can capture and specify the symptoms in detail.

4. Assess whether the questions of the user suggest that specific causes or circumstances lead to certain symptoms.

5. Assess whether the user asks questions in a logical sequence.

6. Assess whether the user reassures the patient that he has received the correct information from the patient.

7. Assess whether the user has summarized his collected information before ending the conversation.

8. Assess whether the user has collected sufficient information of high quality at an appropriate speed.

Assign each of the eight criteria a score according to the following scheme:

1 - Does not meet the criterion  
2 - Rather does not meet the criterion  
3 - Partially meets the criterion  
4 - Rather meets the criterion  
5 - Fully meets the criterion

Explain the evaluation with two sentences.
\normalsize
\end{tcolorbox}

\section{Expert persona prompt}
\label{sec:appendixB}
The expert persona prompt we used is shown below:

\begin{figure*}[h]
    \centering
  \includegraphics[width=0.8\linewidth]{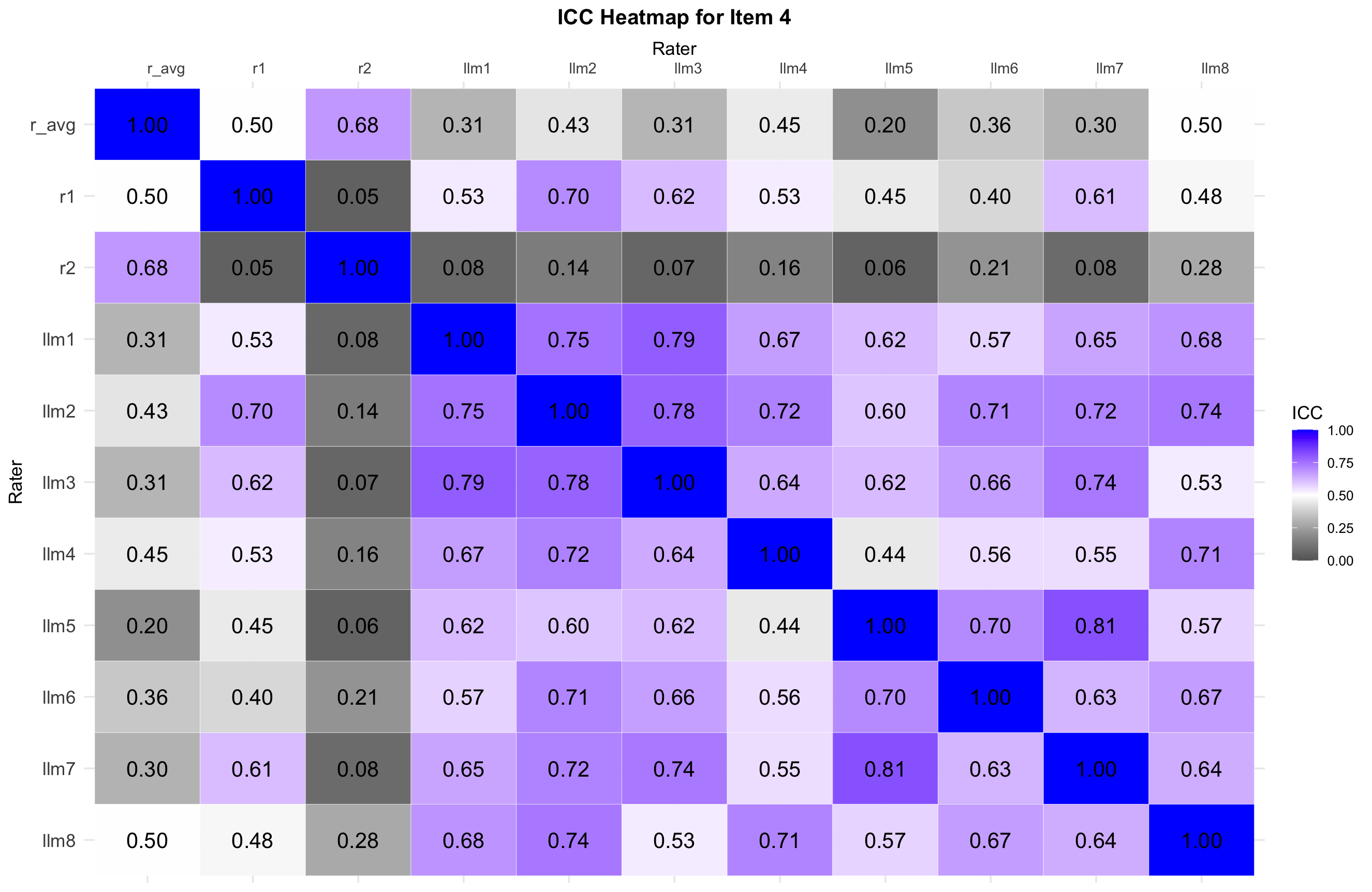}
  \caption{Pairwise Intraclass Correlation Coefficients (ICC) between human and LLM raters across rating conditions for Item 4}
  \label{fig:icc_item4}
\end{figure*}

\begin{tcolorbox}[title=Expert Persona Prompt, colback=gray!5!white, colframe=black!75!white, sharp corners=southwest, fonttitle=\bfseries]
\small
Clinical decision-making (CDM) is a central process in healthcare where physicians gather, evaluate, and interpret relevant information about a patient's health status to make informed decisions regarding diagnosis and treatment. At this point, act as a rater with medical expertise who is evaluating medical students for their CDM mastery. You will assess the quality of the third-semester medical student conducting a history-taking conversation. Provide outputs that a rater with medical expertise would create.

Your assessment should include the following eight criteria
1. Assess whether the user has taken control of the conversation to obtain the necessary information.

2. Assess whether the user recognizes all relevant information.

3. Assess whether the user formulates targeted questions so that he can capture and specify the symptoms in detail.

4. Assess whether the questions of the user suggest that specific causes or circumstances lead to certain symptoms.

5. Assess whether the user asks questions in a logical sequence.

6. Assess whether the user reassures the patient that he has received the correct information from the patient.

7. Assess whether the user has summarized his collected information before ending the conversation.

8. Assess whether the user has collected sufficient information of high quality at an appropriate speed.

Assign each of the eight criteria a score according to the following scheme:

1 - Does not meet the criterion  
2 - Rather does not meet the criterion  
3 - Partially meets the criterion  
4 - Rather meets the criterion  
5 - Fully meets the criterion  

Explain the evaluation with two sentences.
\normalsize
\end{tcolorbox}

\newpage

\section{Pairwise ICC for Item 4 (Figure \ref{fig:icc_item4})}
\label{sec:appendixC}

\section{Main effects \& interaction regression table (Table \ref{tab:fixed-effects})}
\label{sec:appendixD}

\begin{table}[h]
\centering
\footnotesize
\begin{tabular}{@{}l@{\hskip 0.5cm}c@{}}
\toprule
 & Coefficient (SE) \\
\midrule
(Intercept) & 3.523*** (0.070) \\
Large Model & -0.228* (0.096) \\
High Temperature & 0.263** (0.093) \\
Expert Prompt & 0.274** (0.087) \\
Large Model:High Temperature & -0.257* (0.130) \\
Large Model:Expert Prompt & -0.142 (0.132) \\
High Temperature:Expert Prompt & -0.138 (0.121) \\
Large Model:High Temp.:Expert Prompt & 0.149 (0.185) \\
\midrule
Num. Obs. & 189 \\
$R^2$ & 0.293 \\
Adj. $R^2$ & 0.265 \\
AIC & 137.3 \\
BIC & 166.5 \\
RMSE & 0.33 \\
\bottomrule
\end{tabular}
\caption{Fixed effects regression on average clinical reasoning scores, with predictors for model size (Large Model), temperature (High Temperature), and prompt style (Expert Prompt), including interaction terms. Standard errors in parentheses. * $p < .05$, ** $p < .01$, *** $p < .001$.}
\label{tab:fixed-effects}
\end{table}

\end{document}